\documentclass[11pt]{article}

\usepackage[final]{acl} 

\usepackage{times}
\usepackage{latexsym}

\usepackage[T1]{fontenc}

\usepackage[utf8]{inputenc}

\usepackage{microtype}

\usepackage{inconsolata}
\usepackage{graphicx}

\usepackage{subcaption}          
\usepackage{algorithm}
\usepackage{algorithmic}
\usepackage{amsmath,amssymb}
\usepackage{listings}
\usepackage{multirow}
\usepackage{booktabs}
\usepackage[table,xcdraw]{xcolor}
\usepackage{color}
\definecolor{codebackground}{rgb}{0.95,0.95,0.95}
\definecolor{codekeyword}{rgb}{0.13,0.29,0.53}
\definecolor{codecomment}{rgb}{0.13,0.55,0.13}
\definecolor{codestring}{rgb}{0.63,0.125,0.94}
\title{R3-RAG: Learning Step-by-Step Reasoning and Retrieval for LLMs via Reinforcement Learning}
\author{
Yuan Li\thanks{Equal contribution.} \quad
Qi Luo\footnotemark[1] \quad
Xiaonan Li\footnotemark[1] \quad
Bufan Li \quad
Qinyuan Cheng \\
{\bf Bo Wang} \quad
{\bf Yining Zheng} \quad
{\bf Yuxin Wang} \quad
{\bf Zhangyue Yin} \quad
{\bf Xipeng Qiu}\thanks{Corresponding author.} \\
School of Computer Science, Fudan University \\
\texttt{liyuan24@m.fudan.edu.cn} \quad \texttt{xpqiu@fudan.edu.cn}
}

\begin{document}
\maketitle
\begin{abstract}
Retrieval-Augmented Generation (RAG) integrates external knowledge with Large Language Models (LLMs) to enhance factual correctness and mitigate hallucinations. However, dense retrievers often become the bottleneck of RAG systems due to their limited parameters compared to LLMs and their inability to perform step-by-step reasoning. While prompt-based iterative RAG attempts to address these limitations, it is constrained by human-designed workflows.
To address these limitations, we propose \textbf{R3-RAG}, which uses \textbf{R}einforcement learning to make the LLM learn how to \textbf{R}eason and \textbf{R}etrieve step-by-step, thus retrieving comprehensive external knowledge and leading to correct answers. 
R3-RAG is divided into two stages. We first use cold start to make the model learn the manner of iteratively interleaving reasoning and retrieval. 
Then we use reinforcement learning (RL) to further harness its ability to better explore the external retrieval environment.
Specifically, we propose two rewards for R3-RAG: 1) answer correctness for outcome reward, which judges whether the trajectory leads to a correct answer; 2) relevance-based document verification for process reward, encouraging the model to retrieve documents that are relevant to the user question, through which we can let the model learn how to iteratively reason and retrieve relevant documents to get the correct answer.
Experimental results show that R3-RAG significantly outperforms baselines and can transfer well to different retrievers.
\end{abstract}

\section{Introduction}
\begin{figure}[t]
    \centering
    \begin{subfigure}[b]{0.42\textwidth}
        \centering
        \includegraphics[width=\textwidth, ]{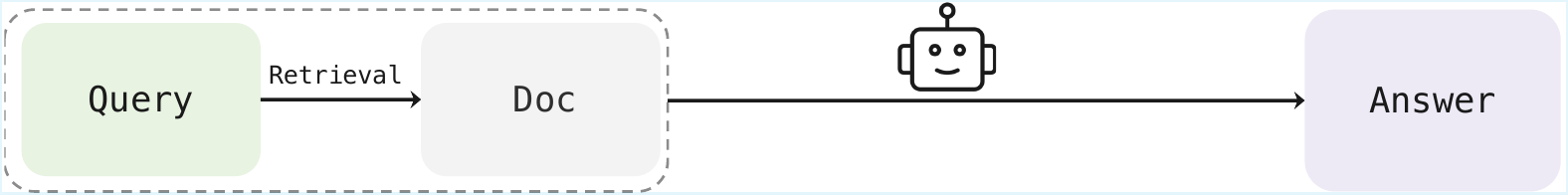}
        \caption{Vanilla RAG}        
        \label{fig:naive-rag}
    \end{subfigure}
    \vspace{0.01\textheight}
    \begin{subfigure}[b]{0.42\textwidth}
        \centering
        \includegraphics[width=\textwidth]{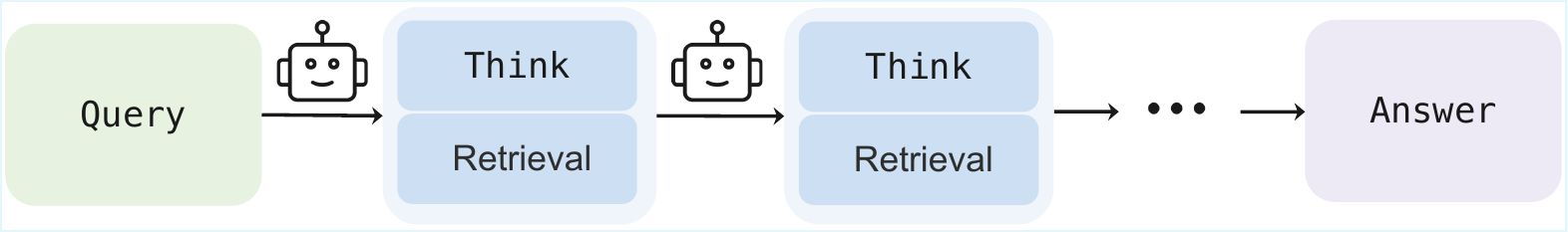}
        \caption{Iterative RAG }
        \label{fig:iterative-rag}
    \end{subfigure}
    \vspace{0.01\textheight}
    \begin{subfigure}[b]{0.42\textwidth}
        \centering
        \includegraphics[width=\textwidth]{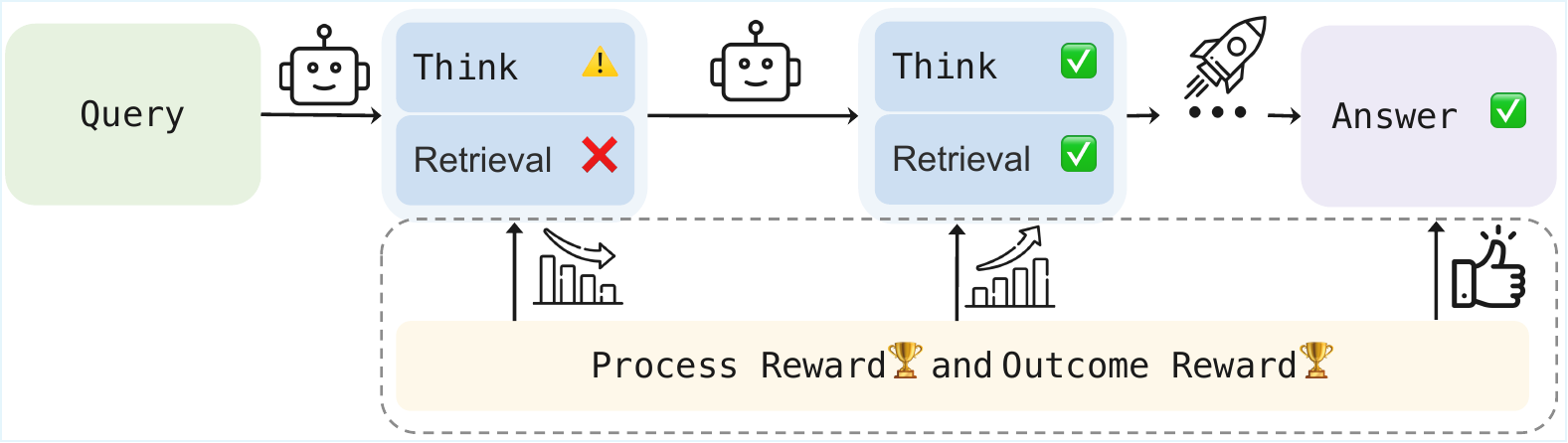}
        \caption{R3-RAG}
        \label{fig:r3-rag}
    \end{subfigure}
    \caption{Comparison of different RAG approaches:
    (a) Vanilla RAG: the LLM uses the documents retrieved for the original question to generate the response; (b) Iterative RAG: the LLM interleaves thinking and invoking the retriever in a fixed, human-designed workflow; and (c) R3-RAG: uses reinforcement learning (RL) to enable the LLM to better reason and retrieve iteratively, to get relevant documents and produce the correct answer.
    }
    \label{fig:rag-comparison}
    \vspace{-3mm}
\end{figure}

Retrieval-Augmented Generation (RAG) has become a prevailing mechanism to address Large Language Models' (LLMs) hallucinations, as shown in Figure~\ref{fig:naive-rag}. It typically first retrieves relevant documents and then integrates them to supplement the LLMs' factual knowledge. In this way, RAG can supplement LLMs with external knowledge, mitigating hallucination and leading to better factuality.

The response quality of RAG highly depends on the retrieval results~\citep{llatrieval, making_llm_robust_to_irrelevant_doc}. 
However, the widely used retrievers have become the bottleneck of the RAG system and have limited the overall performance. Specifically, dense retrievers usually have significantly fewer parameters and have not been scaled to the size of LLMs. Additionally, dense retrievers directly generate vectors for input sequences and cannot conduct step-by-step thinking, which makes it challenging to tackle queries that require reasoning. RAG with iterative retrieval (see Figure~\ref{fig:iterative-rag}) has been proposed to address these limitations. 
For example, \citet{trivedi2023interleavingretrievalchainofthoughtreasoning} propose that LLMs interleave retrieval and Chain-of-Thought (CoT) to better retrieve relevant documents. \citet{llatrieval} propose that LLMs iteratively improve the retrieval results until they are sufficient to support answering users' questions. However, these methods are limited by human-designed workflows and thus cannot fully harness LLMs' reasoning ability. In addition, LLMs are not trained to iteratively invoke retrievers to obtain comprehensive documents, making it challenging for the human-designed workflow to synergize reasoning and retrieval.

To address these limitations, we propose \textbf{R3-RAG}, which uses \textbf{R}einforcement learning (RL) to make the LLM learn how to \textbf{R}eason and \textbf{R}etrieve step-by-step, thus retrieving comprehensive external knowledge and leading to the correct answer. 
As shown in Figure~\ref{fig:r3-rag}, in R3-RAG, the model can explore better reasoning and retrieval trajectories using feedback from the outcome reward and the process reward.
Specifically, R3-RAG is divided into two stages. First, we use a cold start to make the base model learn the manner of iterative reasoning and invoking the retriever. Since a cold start makes it difficult for the model to fully explore the external retrieval environment and thus overshadows the model's ability, we then use RL with the outcome reward to help the model learn how to better reason and retrieve external knowledge step-by-step. To encourage the model to retrieve documents that are relevant to the user's question, we further propose relevance-based document verification as a process reward to measure the relevance of each step's reasoning and retrieval. 
In this way, the model not only learns how to obtain the correct answer through the outcome reward, but also learns to reason and retrieve step-by-step through the fine-grained process reward.

We summarize our contributions as follows:
\vspace{-2mm}
\begin{itemize}
    \item To the best of our knowledge, R3-RAG is the first framework that enables LLMs to perform step-by-step reasoning and retrieval, guided by the outcome reward and the fine-grained process reward.
    \vspace{-3mm}
    \item We conduct comprehensive comparisons with extensive iterative RAG methods, and the results show that R3-RAG significantly outperforms them, indicating its effectiveness.
    \vspace{-3mm}
    \item We will release our code and model to facilitate future research.
\end{itemize}

\section{Related Work}
\paragraph{Retrieval for RAG}
The retrieval in RAG can be mainly divided into two categories: 
1) \textbf{Passive Retrieval:} \citet{realm} implement the Retrieve-Read paradigm, where retrievers fetch documents based on input queries before concatenating them with prompts for LLMs to generate responses, a setup followed by many subsequent works~\citep{lewis2021retrievalaugmentedgenerationknowledgeintensivenlp, alce, udr}. However, dense retrievers cannot perform step-by-step reasoning, and thus, directly using the original question as the query makes it difficult to retrieve comprehensive external information. Iterative retrieval has been proposed to address this limitation; it enables interleaving reasoning and retrieval through decomposed queries.
2) \textbf{Iterative Retrieval:} ReAct, proposed by \citet{yao2023reactsynergizingreasoningacting}, synchronizes reasoning traces with retrieval actions, while Self-Ask, proposed by \citet{press2023measuringnarrowingcompositionalitygap}, implements an autonomous question formulation mechanism during the reasoning process. ITER-RETGEN~\citep{shao2023enhancingretrievalaugmentedlargelanguage} introduces cyclic retrieval-generation patterns; FLARE~\citep{jiang2023activeretrievalaugmentedgeneration} incorporates adaptive retrieval when LLMs generate low-confidence tokens; and IRCoT~\citep{trivedi2023interleavingretrievalchainofthoughtreasoning} strategically positions retrieval operations within CoT reasoning sequences.
Compared with these iterative retrieval methods, which are limited by human-designed workflows, R3-RAG learns to better explore the external retrieval environment via RL, thereby alleviating this limitation.
\vspace{-2mm}

\paragraph{Large Reasoning Models}
Recently, we have seen the emergence of large reasoning models.
\citet{openai2024openaio1card} introduces OpenAI-O1, which demonstrates strong step-by-step reasoning and reflection capabilities. Similarly, \citet{deepseekai2025deepseekr1incentivizingreasoningcapability} develops Deepseek-R1, trained with large-scale RL based on outcome rewards.
\citet{ma2025s2rteachingllmsselfverify} introduces
techniques that enable LLMs to verify their own reasoning steps, leading to more reliable problem-solving capabilities. However, these RL-based methods are highly limited by the base model's knowledge~\citep{base_model_limit_rl}. In this paper, we explore training LLMs to reason and retrieve external information step-by-step, overcoming the base model's knowledge limitations when addressing complex questions.
\citet{guan2025deepragthinkingretrievalstep}, \citet{li2025searcho1agenticsearchenhancedlarge}, \citet{R1-searcher}, and \citet{zheng2025deepresearcherscalingdeepresearch}, concurrent with our work, propose using RL to improve iterative RAG. Our method differs in the critical reward design: while they use only answer correctness as an outcome reward, which cannot provide a fine-grained training signal, we additionally propose relevance-based document verification as a process reward to better guide the model to iteratively retrieve relevant documents and thus obtain comprehensive information for the question.
\begin{figure*}[t]
  \centering
  \includegraphics[width=0.94\textwidth]
  {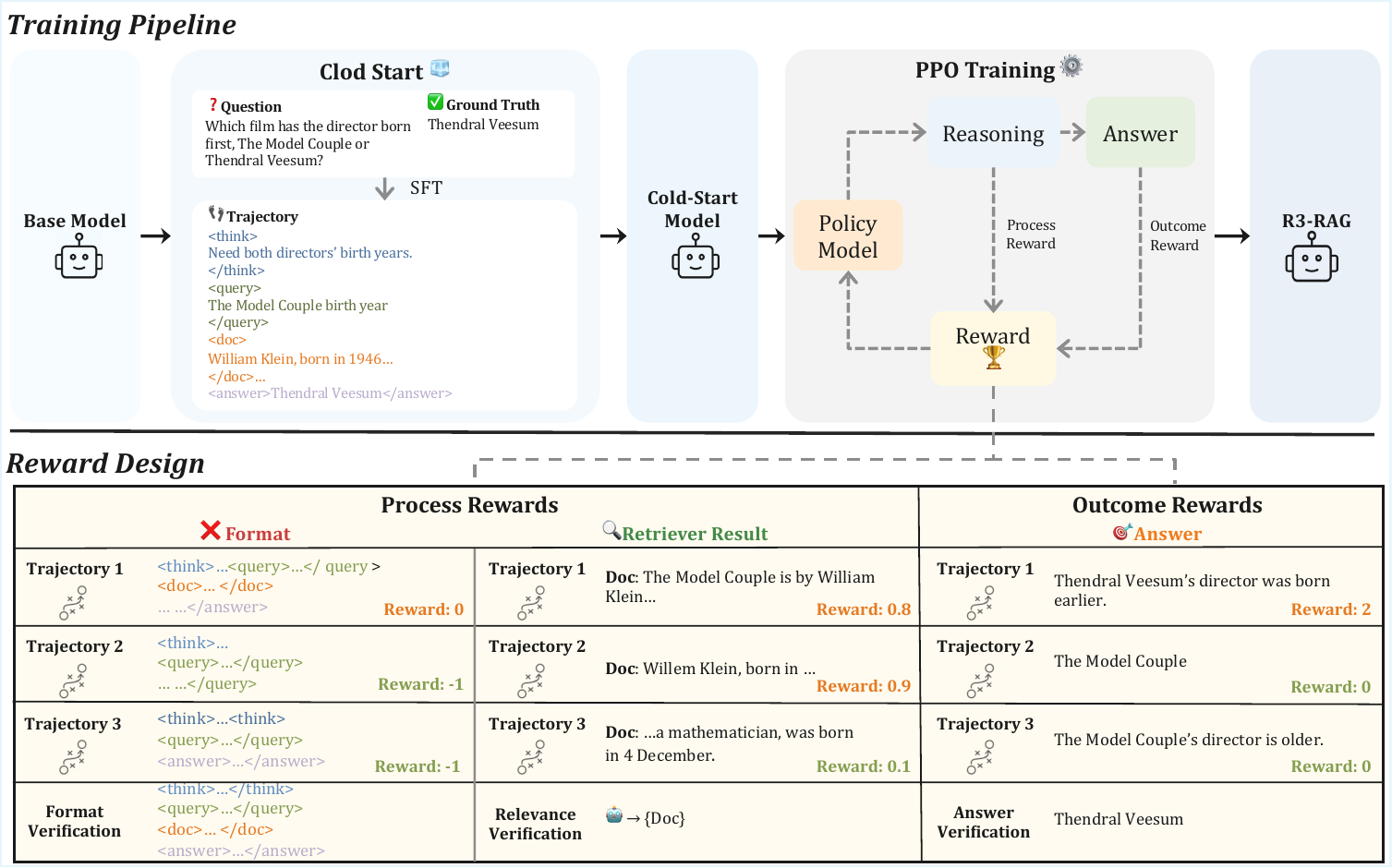}
    \caption{Training Pipeline and Reward Design for R3-RAG}
  \label{fig:main_img}
\end{figure*}

\section{Method}
RAG struggles to generate high-quality responses when retrievers fail to comprehensively retrieve the required knowledge. However, retrievers are usually weaker than LLMs, thus limiting the LLM's ability. To address this limitation, we propose \textbf{R3-RAG}, which uses \textbf{R}einforcement learning to enable the LLM to learn to \textbf{R}eason and \textbf{R}etrieve step-by-step, thereby retrieving comprehensive external knowledge and generating a factual answer. Specifically, R3-RAG is divided into two stages, as shown in Figure~\ref{fig:main_img}. 
First, we leverage cold start~\citep{deepseekai2025deepseekr1incentivizingreasoningcapability} to teach the base model the manner of iterative reasoning and invoking the retriever. 
Since a cold start makes it difficult for the model to fully explore the external retrieval environment, we then use RL to further harness its ability to better search for relevant external knowledge.
Specifically, we propose two rewards for R3-RAG: 1) an answer-correctness outcome reward, which judges whether the trajectory leads to a correct answer against the ground truth; and 2) relevance-based document verification as a process reward, encouraging the model to retrieve documents that are relevant to the user's question.
In this section, we first introduce the R3-RAG trajectory and then the cold start and RL training.

\subsection{Trajectory Definition}
In this section, we define the R3-RAG trajectory.
Given a multi-hop question-answering (QA) dataset $\mathcal{Q} = \{(q_i, a_i^*)\}_{i=1}^{n}$, where $q_i$ is a question and $a_i^*$ is its ground-truth answer, we define a trajectory $T_i$ for each question $q_i$ as a sequence of reasoning and retrieval steps $T_i = \{s_i^1, s_i^2, \ldots, s_i^{|T_i|}\}$. 
Each step $s_i^j$ in trajectory $T_i$ is as follows:

\begin{equation}
\small
s_i^{j} = 
\begin{cases}
(\xi_i^{j},\,\psi_i^{j},\,d_i^{j}) & j < \lvert T_i\rvert\\[4pt]
(\xi_i^{j},\,a_i)                   & j = \lvert T_i\rvert
\end{cases},
\end{equation}
where the model iteratively reasons and invokes the retriever until it deems the documents sufficient to generate the answer at step $\lvert T_i\rvert$.
$\boldsymbol{\xi_i^{j}}$ denotes the \textbf{Reasoning Process} at step $j$, which analyzes all information (e.g., analysis and documents) accumulated up to step $j-1$, and plans the action for step $j$;
$\boldsymbol{\psi_i^{j}}$ is the \textbf{Retrieval Query} produced by this reasoning;  
$\boldsymbol{d_i^{j}}$ comprises the \textbf{Documents} returned by the retriever for that query;  
and $\boldsymbol{a_i}$ is the final \textbf{Answer} to the original question.

During the reasoning process $\boldsymbol{\xi_i^{j}}$, the model can perform step-by-step analysis of the information gathered in previous steps.  
If the existing documents are sufficient to answer the question, the model outputs the answer $\boldsymbol{a_i}$; otherwise, it determines what additional information is needed and invokes the retriever with a new query $\boldsymbol{\psi_i^{j}}$.

\subsection{Cold Start}
Base models do not natively interleave step-by-step reasoning with retrieval. We therefore begin with a cold-start stage that synthesizes supervised trajectories to teach the desired interleaved behavior.

\begin{algorithm}[t] 
\small
\caption{Cold-Start Data Generation}
\label{alg:reasoning-retrieval}
\begin{algorithmic}[1]
\REQUIRE Dataset $\mathcal{Q}_{CS} = \{(q_i, a_i^*)\}_{i=1}^{n}$, Retriever $R$, Language Model $M$
\ENSURE Trajectory-Augmented Dataset $\mathcal{D}_{CS}$
\STATE $\mathtt{MAX\_ITERATION}  \gets 5 $, $\mathcal{D}_{CS}   \gets \emptyset$\\

\FOR{each $(q_i, a_i^*) \in \mathcal{Q}_{CS}$}
    \STATE $T_i \gets [\,]$
    \FOR{turn $\in [1, \mathtt{MAX\_ITERATION}]$}
        \STATE $\mathtt{step} \gets \mathtt{Parse}(M(\mathtt{Prompt}(q_i, T_i)))$
        \IF{$\texttt{answer} \in \mathtt{step}$}
            \STATE $T_i.\mathtt{append}(\mathtt{step})$
            \STATE \textbf{break}
        \ELSIF{$\texttt{query} \in \mathtt{step}$}
            \STATE $\mathtt{step}[\texttt{doc}] \gets R(\mathtt{step}[\texttt{query}])$
            \STATE $T_i.\mathtt{append}(\mathtt{step})$
            \IF{$\mathtt{turn} == \mathtt{MAX\_ITERATION}$}
                \STATE \textbf{break} \textit{Fail to generate the answer}
            \ENDIF
        \ELSE
            \STATE \textbf{break} \textit{The format of $T_i$ is invalid}
        \ENDIF
    \ENDFOR
    \STATE $\mathcal{D}_{CS} \gets \mathcal{D}_{CS} \cup \{(q_i, a_i^*, T_i)\}$
\ENDFOR
\RETURN $\mathcal{D}_{CS}$
\end{algorithmic}
\end{algorithm}

\subsubsection{Cold-Start Data Generation}
Algorithm~\ref{alg:reasoning-retrieval} outlines the procedure. For each question, we iteratively construct a trajectory whose steps follow a fixed schema. At each iteration, a powerful LLM is prompted to produce 1) a reasoning step with a retrieval query, or 2) a reasoning step with a final answer. For the first step, the template in Figure~\ref{fig:template1} prompts the model to analyze the problem, outline a solution path, and decide whether to retrieve or answer directly. For subsequent steps, the template in Figure~\ref{fig:template2} asks the model to consolidate prior reasoning and retrieved documents, assess the sufficiency of evidence, identify remaining knowledge gaps, and decide whether further retrieval is necessary or a final answer can be produced. If a final answer is produced, trajectory construction terminates. The resulting instance pairs each question with an interleaved reasoning–retrieval trajectory.

To improve data quality, we vary the sampling temperature and generate multiple trajectories per question, then perform rejection sampling against the ground-truth answer, discarding trajectories whose final answer is incorrect or whose steps violate the required format. The remaining trajectories constitute the cold-start dataset $\mathcal{D}_{\mathrm{CS}}$.

\subsubsection{Cold-Start Training}
To use $\mathcal{D}_{CS}$ for training in the cold-start stage—i.e., supervised fine-tuning (SFT)—we design Algorithm~\ref{alg:training-data-generation} to convert $\mathcal{D}_{CS}$ into an SFT-ready dataset $\mathcal{D}_{\mathrm{CS}}^{\mathrm{train}} = \{(\mathtt{input}_k, \mathtt{output}_k)\}_{k=1}^{m}$, where $m = \sum_{i=1}^{|\mathcal{Q}_{CS}|} |T_i|$. 
For each tuple $(q_i,a_i^*,T_i)\in\mathcal{D}_{CS}$, we generate $|T_i|$ input–output pairs.
Consider a step $s_i^{j}$ in trajectory $T_i$.
We construct the input by concatenating the question $q_i$ with all steps that \emph{precede} $s_i^{j}$, serialized into our unified format, including the reasoning (analysis $\xi_i^{j'}$ and retrieval query $\psi_i^{j'}$) and the retrieved documents $d_i^{j'}$ for all $j'<j$.
We then construct the output as the reasoning content of the current step $s_i^{j}$ serialized in the same format.
Iterating over all steps in $T_i$ yields $|T_i|$ training instances, which we add to $\mathcal{D}_{\mathrm{CS}}^{\mathrm{train}}$.

\begin{algorithm}[t]
\small
\caption{Training Data Generation for SFT}
\label{alg:training-data-generation}
\begin{algorithmic}[1]
\REQUIRE Trajectory-augmented dataset $\mathcal{D}_{CS}$
\ENSURE SFT training dataset $\mathcal{D}_{\mathrm{CS}}^{\mathrm{train}}$
\STATE $\mathcal{D}_{\mathrm{CS}}^{\mathrm{train}} \gets \emptyset$
\FOR{each $(q_i, a_i^*, T_i) \in \mathcal{D}_{CS}$}
    \STATE $\mathtt{context} \gets q_i$
    \STATE $\mathtt{history} \gets \emptyset$
    \FOR{$j = 1$ \TO $|T_i|$}
        \STATE $\mathtt{input} \gets \mathtt{Concat}(\mathtt{context}, \mathtt{history})$
        \STATE $s_i^j \gets T_i[j]$
        \IF{$j < |T_i|$}
            \STATE $(\xi_i^{j},\,\psi_i^{j},\,d_i^{j}) \gets s_i^j$
            \STATE $\mathtt{output} \gets \mathtt{Format}(\xi_i^{j},\,\psi_i^{j})$
            \STATE $\mathtt{step} \gets \mathtt{Format}(\xi_i^{j},\,\psi_i^{j},\,d_i^{j})$
            \STATE $\mathtt{history} \gets \mathtt{Concat}(\mathtt{history}, \mathtt{step})$
        \ELSE
            \STATE $(\xi_i^{j},\,a_i) \gets s_i^j$
            \STATE $\mathtt{output} \gets \mathtt{Format}(\xi_i^{j},\,a_i)$
        \ENDIF
        \STATE $\mathtt{pair} \gets (\mathtt{input}, \mathtt{output})$
        \STATE $\mathcal{D}_{\mathrm{CS}}^{\mathrm{train}} \gets \mathcal{D}_{\mathrm{CS}}^{\mathrm{train}} \cup \{\mathtt{pair}\}$
    \ENDFOR
\ENDFOR
\RETURN $\mathcal{D}_{\mathrm{CS}}^{\mathrm{train}}$
\end{algorithmic}
\end{algorithm}

Based on $\mathcal{D}_{\mathrm{CS}}^{\mathrm{train}}$, we perform SFT with the maximum likelihood estimation objective to obtain the R3-RAG-CS model.
This initialization enables the model to iteratively reason and retrieve, which can help the model better sample positive trajectories and thus yield stronger RL training signals.

\subsection{Reinforcement Learning}
Although the cold start can enable the model to learn the manner of iterative reasoning and retrieval, it is hard to make the model fully explore the external retrieval environment, thus leaving it limited by the cold-start data. To address these challenges, we use RL to further harness the model's own abilities to better reason and retrieve external information. Specifically, we propose two rewards for R3-RAG: 1) answer correctness as an outcome reward, which judges whether the trajectory leads to correct answers by ground truth; and 2) relevance-based document verification as a process reward, encouraging the model to retrieve documents that are highly relevant to users' questions. In this section, we first introduce the reward design and then the training objective.

\subsubsection{Reward Design}
\paragraph{Answer Correctness}
The high-quality outcome reward is critical for the model to generate the correct trajectory~\citep{deepseekai2025deepseekr1incentivizingreasoningcapability}. For question-answering, string matching is usually used to judge whether the prediction is correct according to the ground-truth answer. However, since the same entity can have different variants, the match-based judgment can be inaccurate, which introduces extra noise to the training signal. We propose to combine the match-based judgment and the model-based judgment as follows:

\begin{small}
\begin{align}
    \mathrm{Acc}(a) = 
\begin{cases}
1 & \text{if } \mathrm{Acc}_{match}(a)  \operatorname{or}  \mathrm{Acc}_{model}(a)\\
  0 & otherwise
\end{cases},
\end{align}
\end{small}

\noindent where $a$ is the predicted answer, with $\mathrm{Acc}_{\text{match}}$ and $\mathrm{Acc}_{\text{model}}$ denoting the match-based and model-based answer judgments, respectively.

\paragraph{Document Relevance}
Using only the outcome reward is challenging because it provides a limited fine-grained training signal and results in sparse rewards, which degrades RL effectiveness. We further propose relevance-based document verification to give fine-grained guidance to the process of iterative retrieval. In each retrieval step, we calculate the document relevance as:

\begin{small}
\begin{align}
    \mathrm{Rel(d_i^j)} = \mathrm{LLM} (I_{relevance}, q_i, d_i^j),
\end{align}
\end{small}

\noindent where $I_{relevance}$ is the instruction for the LLM to judge the relevance between the question $q_i$ and the document $d_i^j$, with scores ranging from $0, 0.1, 0.2, \dots$ to $1$.
Since $q_i$ and $d_i^j$ are concatenated and then sent to the LLM, the relevance judgment can take advantage of full token-level interaction and thus guide the model to better retrieve relevant information.

\paragraph{Format Correctness}
We also implement a format reward $Val(s_i^j)$ to encourage the model to response in the required format as follows:

\begin{small}
\begin{align}
    \mathrm{Val(s_i^j)} = 
\begin{cases}
1 & \text{if } s_i^j \text{ is valid}\\
0 & otherwise
\end{cases}.
\end{align}
\end{small}

\paragraph{Overall Reward}
The overall reward $r(s_i^j)$ is defined as:
\begin{equation}
\small
r(s_i^j) = Val(s_i^j) \cdot (\mathrm{Acc}(a_i) + \mathrm{Rel}(d_i^j)) + Val(s_i^j) - 1,
\end{equation}
where if $s_i^j$ denotes the final step, $a_i$ represents the predicted answer, and $\mathrm{Rel}(d_i^j)$ is set to 0 since no retrieved document $d_i^j$ is involved in this step. Otherwise, if $s_i^j$ corresponds to an intermediate step, $d_i^j$ indicates the retrieved documents at that step, and $\mathrm{Acc}(a_i)$ is set to 0 as no answer is available.

In this way, we can teach the model to iteratively invoke the retriever and obtain external information to achieve the correct final answer.
To enhance the effect of answer-correctness-based reward and prevent the training from being distracted by other rewards, we use the following factor to adjust the process reward throughout the trajectory:
\begin{equation}
\small
r_{all}(s_i^j) = r(s_i^j) \cdot \lambda_T(T_i),
\end{equation}
where $\lambda_T(T_i)$ is determined based on the trajectory. For trajectories with correct answers, set $\lambda_T(T_i) > 1$ to reinforce successful reasoning; for incorrect answers, set $0 < \lambda_T(T_i) < 1$ to down-weight the process reward; for trajectories with format errors, set $\lambda_T(T_i) = 1$ to isolate format penalties to specific steps.

\begin{table*}[t]
\small
\centering
\begin{tabular}{@{}lccccccccccc@{}}
\toprule
\multicolumn{1}{c}{}                                   &                                      & \multicolumn{3}{c}{\textbf{HotpotQA}}         & \multicolumn{3}{c}{\textbf{2WikiMultiHopQA}}  & \multicolumn{3}{c}{\textbf{MuSiQue}}          & \textbf{Average}  \\ \cmidrule(l){3-12} 
\multicolumn{1}{c}{\multirow{-2}{*}{\textbf{Methods}}} & \multirow{-2}{*}{\textbf{Retriever}} & ACC           & F1            & EM            & ACC           & F1            & EM            & ACC           & F1            & EM            & ACC           \\ \midrule
\multicolumn{12}{c}{\cellcolor[HTML]{EFEFEF}\textbf{Llama-3.1-8B}}                                                                                                                                                                                            \\
CoT                                                    & -                                    & 39.2          & 38.3          & 27.6          & 28.8          & 34.5          & 25.3          & 14.0          & 17.2          & 8.0           & 27.3          \\
RAG with CoT                                           & E5                                   & 53.3          & 50.8          & 38.6          & 32.9          & 40.2          & 31.1          & 16.3          & 18.8          & 10.3          & 34.2          \\
ReAct(Tool Call)                                       & E5                                   & 30.8          & 25.8          & 21.2          & 17.8          & 14.5          & 12.9          & 8.6           & 6.5           & 4.8           & 19.1          \\
Flare                                                  & E5                                   & 24.6          & 20.9          & 17.8          & 11.9          & 11.4          & 10.9          & 3.5           & 2.8           & 2.3           & 13.3          \\
Self-ask                                               & E5                                   & 39.1          & 34.3          & 29.5          & 29.0          & 27.5          & 26.2          & 14.4          & 23.5          & 13.5          & 27.5          \\
ITER-RETGEN                                            & E5                                   & 45.7          & 40.0          & 34.3          & 19.4          & 18.1          & 17.0          & 11.1          & 8.9           & 7.9           & 25.4          \\
IRCoT                                                  & E5                                   & 52.8          & 46.0          & 39.3          & 40.6          & 37.5          & 35.1          & 16.7          & 13.6          & 12.0          & 36.7          \\
Auto-RAG                                               & BM25                                 & -             & 36.1          & 25.8          & -             & 30.1          & 23.0          & -             & -             & -             & -             \\
DeepRAG                                                & BM25                                 & -             & 51.5          & 40.7          & -             & 53.3          & 48.1          & -             & -             & -             & -             \\
\midrule[0.03em]
R3-RAG-CS                                              & E5                                   & 60.6          & 55.5          & 43.6          & 53.7          & 53.4          & 46.5          & 29.6          & 29.4          & 19.4          & 48.0          \\
R3-RAG                                                 & E5                                   & 64.4          & 58.8          & 45.6          & 61.0          & 60.9          & 52.9          & 32.2          & 32.7          & 21.1          & 52.6          \\
R3-RAG                                                 & BM25                                 & 62.5          & 57.6          & 44.4          & 58.0          & 58.6          & 50.6          & 26.4          & 27.7          & 17.2          & 49.0          \\
R3-RAG                                                 & BGE                                  & \textbf{65.3} & \textbf{60.0} & \textbf{46.6} & \textbf{62.1} & \textbf{61.8} & \textbf{53.7} & \textbf{33.8} & \textbf{32.8} & \textbf{21.2} & \textbf{53.8} \\ \midrule
\multicolumn{12}{c}{\cellcolor[HTML]{EFEFEF}\textbf{Qwen2.5-7B}}                                                                                                                                                                                              \\
CoT                                                    & -                                    & 34.0          & 34.0          & 23.0          & 31.1          & 36.9          & 27.4          & 12.7          & 17.0          & 6.7           & 25.9          \\
RAG with CoT                                           & E5                                   & 52.4          & 49.4          & 37.6          & 33.5          & 39.1          & 30.1          & 16.9          & 18.8          & 9.9           & 34.3          \\
ReAct(Tool Call)                                       & E5                                   & 27.9          & 23.1          & 21.1          & 38.3          & 31.7          & 26.1          & 11.7          & 8.0           & 5.8           & 26.0          \\
Flare                                                  & E5                                   & 28.9          & 24.8          & 21.2          & 29.1          & 28.3          & 27.8          & 6.5           & 4.8           & 3.6           & 21.5          \\
Self-ask                                               & E5                                   & 37.1          & 33.2          & 28.7          & 31.5          & 29.8          & 28.2          & 13.3          & 10.7          & 8.6           & 27.3          \\
ITER-RETGEN                                            & E5                                   & 48.4          & 40.2          & 35.8          & 32.8          & 31.4          & 30.2          & 12.3          & 10.1          & 8.5           & 31.2          \\
IRCoT                                                  & E5                                   & 48.4          & 41.1          & 35.7          & 35.8          & 33.5          & 31.1          & 13.5          & 11.2          & 9.4           & 32.6          \\
DeepRAG                                                & BM25                                 & -             & 41.1          & 32.1          & -             & 44.9          & 40.4          & -             & -             & -             & -             \\
ReSearch-7B                                            & BGE                                  & 60.3          & -             & 40.6          & 50.1          & -             & 44.7          & 32.2          & -             & 21.7          & 47.5          \\
ReSearch-7B-Instruct                                   & BGE                                  & 63.6          & -             & 43.5          & 54.2          & -             & 47.6          & 33.4          & -             & 22.3          & 50.4          \\
\midrule[0.03em]
R3-RAG-CS                                              & E5                                   & 63.3          & 57.6          & 45.2          & 53.1          & 53.5          & 46.1          & 31.3          & 31.9          & 21.9          & 49.2          \\
R3-RAG                                                 & E5                                   & 65.5          & 59.7          & 46.4          & 62.3          & 62.7          & 54.2          & 33.6          & 34.0          & 21.4          & 53.8          \\
R3-RAG                                                 & BM25                                 & 63.8          & 58.2          & 44.9          & 59.6          & 61.1          & 52.8          & 29.2          & 30.0          & 17.6          & 50.8          \\
R3-RAG                                                 & BGE                                  & \textbf{66.4} & \textbf{60.6} & \textbf{46.8} & \textbf{63.0} & \textbf{63.4} & \textbf{55.2} & \textbf{34.8} & \textbf{34.3} & \textbf{21.7} & \textbf{54.8} \\ \bottomrule
\end{tabular}
\caption{
Performance comparison of different models and retrieval methods across multi-hop QA datasets. Our R3-RAG method consistently outperforms all baselines across accuracy (ACC), F1, and exact match (EM) metrics, regardless of the base model or retriever used. Bold values indicate the best performance in each column and model group. Missing values (-) indicate results not reported in the original papers.
\vspace{-10pt}
}
\label{tab:main-results}
\end{table*}

\subsubsection{Training Objective}
Guided by the outcome and process reward, we apply the Proximal Policy Optimization (PPO) to train R3-RAG-CS model:

{
    \small
    \begin{align}
    \mathcal{L}_{RL} =& \mathbb{E}_{(x,y)\sim \mathcal{D}_{\mathrm{RL}}^{\mathrm{train}}} \Big[ \min\Big(\rho(\theta) A(x,y), \nonumber \\
    &\operatorname{clip}(\rho(\theta), 1-\epsilon, 1+\epsilon) A(x,y) \Big) \Big] \\
    &+ \beta \mathcal{L}_{KL}, \nonumber
    \end{align}
}

\noindent where $\rho(\theta) = \frac{P(y|x;\theta)}{P(y|x;\theta_{old})}$ represents the probability ratio between the current policy and the old policy, $A(x,y)$ denotes the advantage function estimated using Generalized Advantage Estimation (GAE), $\epsilon$ is the clipping parameter that constrains policy updates, $\mathcal{L}_{KL}$ represents the Kullback-Leibler ({KL}) divergence between the current policy and the reference model to prevent excessive deviation, and $\beta$ controls the strength of the KL penalty~\citep{ppo}. Details are provided in Appendix~\ref{app:gae}.

\section{Experiments}
\subsection{Datasets and Metrics}
We follow \citet{R1-searcher} to conduct a comparison on three multi-hop QA datasets: HotpotQA \cite{yang2018hotpotqadatasetdiverseexplainable}, 2WikiMultiHopQA \cite{ho2020constructingmultihopqadataset}, and MuSiQue \cite{trivedi2022musiquemultihopquestionssinglehop}, using their full dev sets. 
In the RL stage, we use only 8,192 examples from HotpotQA for training.
We evaluate using three complementary metrics: \textbf{Exact Match} (EM) measures strict string matching against ground-truth answers; \textbf{F1 score} quantifies partial matching via token overlap between predicted and ground-truth answers; and \textbf{Accuracy} (Acc) is based on LLM judgment of semantic correctness. For EM and F1, we first extract the short answer from the LLM's response and then compare it against the ground-truth answer.

\subsection{Baselines}
We compare against several competitive baselines: 1) \textbf{CoT}: LLMs directly answer the question using CoT; 
2) \textbf{RAG with CoT}: LLMs answer the question based on documents retrieved using original questions as queries;
3) Iterative retrieval methods: LLMs are prompted to invoke retrieval tools during reasoning, including \textbf{ReAct} \cite{yao2023reactsynergizingreasoningacting}, \textbf{FLARE} \cite{jiang2023activeretrievalaugmentedgeneration}, \textbf{Self-Ask} \cite{press2023measuringnarrowingcompositionalitygap}, \textbf{ITER-RETGEN} \cite{shao2023enhancingretrievalaugmentedlargelanguage}, and \textbf{IRCoT} \cite{trivedi2023interleavingretrievalchainofthoughtreasoning}; 
and 4) Fine-tuned methods, including \textbf{Auto-RAG} \cite{yu2024autoragautonomousretrievalaugmentedgeneration}, \textbf{DeepRAG} \cite{guan2025deepragthinkingretrievalstep}, and \textbf{ReSearch} \cite{chen2025researchlearningreasonsearch}.

\subsection{Implementation Details}
For the cold-start stage, we create 51,254 solution trajectories from the training sets of HotpotQA~\cite{yang2018hotpotqadatasetdiverseexplainable}, 2WikiMultiHopQA~\cite{ho2020constructingmultihopqadataset}, and MuSiQue~\cite{trivedi2022musiquemultihopquestionssinglehop}. For the RL stage, we sample 8,192 instances from the remaining HotpotQA training data.
We employ E5-base-v2~\citep{wang2022text} as our default retriever. During evaluation, the models receive complete retrieved documents, limited to the top 5 per iteration with a maximum of 5 iterations. Unlike prior work~\citep{zheng2025deepresearcherscalingdeepresearch}, which extracts information from retrieved documents to reduce context length, our approach directly incorporates complete retrieved documents.
All methods are evaluated using the same metrics and hyperparameters. Further details are provided in Appendix~\ref{app:details}.

\subsection{Main Results}
We show the evaluation results of R3-RAG using Llama-3.1-8B and Qwen2.5-7B in Table~\ref{tab:main-results}. We see that R3-RAG significantly outperforms baselines on three datasets across different metrics, which indicates strong overall reasoning and retrieval performance. Specifically, R3-RAG outperforms the strongest iterative RAG workflow method, IRCoT, by around 15 percentage points on average, which shows that R3-RAG can help the model better learn to iteratively reason and retrieve step-by-step and thus get more accurate answers.
Additionally, R3-RAG also outperforms ReSearch~\citep{chen2025researchlearningreasonsearch}, which uses RL with an outcome-based reward to augment the model's iterative retrieval ability. This shows the advantages of R3-RAG's reward design, which not only teaches the model to get the final answer but also guides the model to retrieve documents relevant to the question.
Meanwhile, R3-RAG achieves significant improvements with both Qwen2.5-7B and Llama-3.1-8B, and this shows the robustness of R3-RAG. Although R3-RAG's RL is only trained on the HotpotQA dataset, R3-RAG outperforms the cold-start model on three datasets, which shows that the proposed RL algorithm can generalize well across various datasets and further demonstrates R3-RAG's transferability to different downstream scenarios.

\section{Analysis}
\subsection{Effect of Different Reward}  
Table~\ref{tab:rewardmodel-results} presents the comparative performance of different reinforcement learning configurations in R3-RAG. Our full model employs both process and outcome rewards. When we remove the process reward component from our complete R3-RAG framework, we observe a performance decline of 1.4 percentage points on average (from 46.6\% to 45.2\%), which shows that fine-grained process rewards effectively guide the model to retrieve more relevant documents at each reasoning step, leading to improved overall performance. Further removing the outcome reward component (thus eliminating all reinforcement learning signals) results in an additional performance drop of 3.6 percentage points on average (from 45.2\% to 41.6\%).  These results demonstrate the effectiveness of both reward components. Outcome rewards guide the model to optimize for correct final answers, while process rewards provide granular feedback on the quality of each retrieval operation. By jointly optimizing for both the ultimate goal (correct answers) and the intermediate steps (relevant document retrieval), our approach achieves a more effective reasoning-retrieval strategy.
\begin{table}[t]
\small
\centering
\begin{tabular}{@{}
>{\columncolor[HTML]{FFFFFF}}l 
>{\columncolor[HTML]{FFFFFF}}c 
>{\columncolor[HTML]{FFFFFF}}c 
>{\columncolor[HTML]{FFFFFF}}c @{}}
\toprule
\multicolumn{1}{c}{\cellcolor[HTML]{FFFFFF}\textbf{Methods}} & \textbf{2Wiki} & \textbf{MuSiQue} & \textbf{Avg}  \\ \midrule
R3-RAG                                                       & \textbf{61.0}          & \textbf{32.2}    & \textbf{46.6} \\
~~~- Process Reward                                             & 59.0                   & 31.4             & 45.2          \\
~~~~~~ - Outcome Reward                                                         & 53.7                   & 29.6             & 41.6          \\ \bottomrule
\end{tabular}

\caption{Effect of Different Reward on R3-RAG.}
\label{tab:rewardmodel-results}
\end{table}

\begin{figure}[b]
    \centering
    \includegraphics[width=0.49\textwidth]{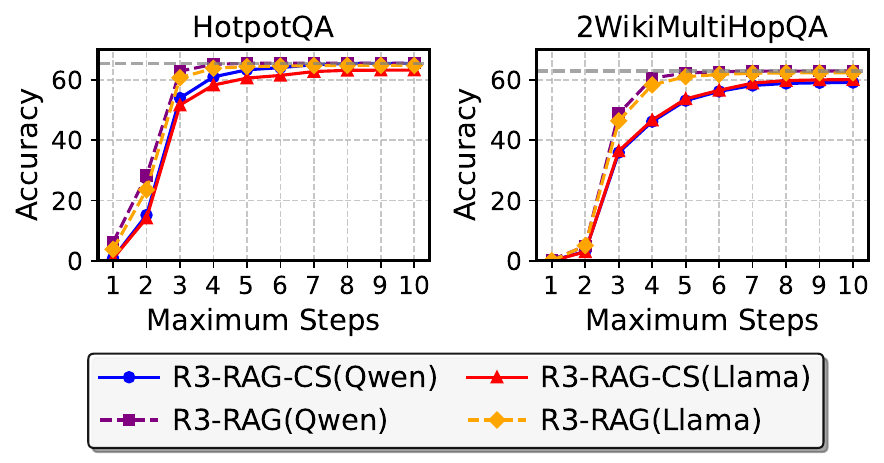}
    \caption{Impact of the maximum number of reasoning steps on HotpotQA and 2WikiMultiHopQA. Results are shown for both Qwen and Llama backbone models. All models are trained with up to five reasoning steps.}
    \label{fig:depth_main}
\end{figure}

\subsection{Effect of Maximum Iteration Steps}
In this section, we analyze how the number of reasoning steps affects model performance. As shown in Figure~\ref{fig:depth_main}, R3-RAG demonstrates steady improvements in answer accuracy as the number of steps increases from 1 to 10. Notably, R3-RAG continues to benefit from additional reasoning steps at inference time, even beyond the maximum number used during training (five steps). This suggests that our framework enables models to generalize their reasoning procedures effectively to longer inference trajectories.
Furthermore, we observe that R3-RAG consistently outperforms R3-RAG-CS when evaluated with the same maximum number of steps, highlighting the effectiveness of our RL training approach in enabling models to more accurately answer questions while using the same number of reasoning steps.

\subsection{Impact of Retrieval Top-$k$ Parameter}  
\begin{figure}[t]
    \centering
    \includegraphics[width=0.49\textwidth]{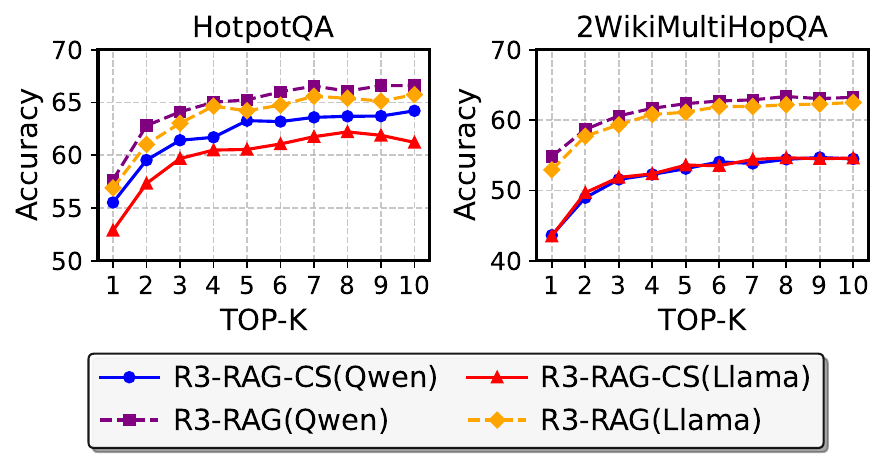}
    \caption{R3-RAG performance across different retrieval top $k$.}
    \label{fig:topk_main}
\end{figure}

We analyze how the number of documents retrieved at each step affects our method’s performance. Figure~\ref{fig:topk_main} demonstrates that increasing top $k$ from 1 to 5 significantly improves performance by approximately 8 percentage points, while adding more documents beyond 5 yields minimal gains (less than 1\%). Additionally, we observe that our RL approach substantially enhances answer accuracy across all top-$k$ configurations, demonstrating the robustness of our RL method regardless of variation in the number of retrieved documents per step.

\subsection{Transferability over Different Retrievers}  
\begin{table}[t]
\small
\centering
\begin{tabular}{@{}
>{\columncolor[HTML]{FFFFFF}}l 
>{\columncolor[HTML]{FFFFFF}}c 
>{\columncolor[HTML]{FFFFFF}}c 
>{\columncolor[HTML]{FFFFFF}}c 
>{\columncolor[HTML]{FFFFFF}}c @{}}
\toprule
\multicolumn{1}{c}{\cellcolor[HTML]{FFFFFF}\textbf{Methods}} & \textbf{Retriever} & \textbf{Hotpot} & \textbf{2Wiki} & \textbf{Avg} \\ \midrule
\multicolumn{5}{c}{\cellcolor[HTML]{EFEFEF}\textbf{Llama-3.1-8B}}                                                                   \\
CoT with RAG                                                 & BM25               & 52.3            & 31.5           & 41.9         \\
\textbf{}                                                    & E5                 & 53.3            & 32.9           & 43.1         \\
\textbf{}                                                    & BGE                & 55.8            & 33.9           & 44.8         \\
ReAct                                                        & BM25               & 33.6            & 18.0           & 25.8         \\
                                                             & E5                 & 30.8            & 17.8           & 24.3         \\
                                                             & BGE                & 29.8            & 16.1           & 22.9         \\
IRCoT                                                        & BM25               & 52.8            & 41.4           & 47.1         \\
                                                             & E5                 & 52.8            & 40.6           & 46.7         \\
                                                             & BGE                & 43.3            & 20.8           & 32.0         \\
\midrule[0.03em]
R3-RAG                                                       & BM25               & 62.5            & 58.0           & 60.2         \\
\textbf{}                                                    & E5                 & 64.4            & 61.0           & 62.7         \\
\textbf{}                                                    & BGE                & 65.3            & 62.1           & 63.7         \\ \midrule
\multicolumn{5}{c}{\cellcolor[HTML]{EFEFEF}\textbf{Qwen2.5-7B}}                                                                     \\
CoT with RAG                                                 & BM25               & 49.9            & 29.8           & 39.9         \\
                                                             & E5                 & 52.4            & 33.5           & 42.9         \\
                                                             & BGE                & 55.1            & 35.3           & 45.2         \\
ReAct                                                        & BM25               & 43.4            & 28.7           & 36.1         \\
                                                             & E5                 & 27.9            & 38.3           & 33.1         \\
                                                             & BGE                & 37.7            & 27.4           & 32.6         \\
IRCoT                                                        & BM25               & 48.5            & 33.8           & 41.2         \\
                                                             & E5                 & 48.4            & 35.8           & 42.1         \\
                                                             & BGE                & 40.5            & 20.1           & 30.3         \\
\midrule[0.03em]
R3-RAG                                                       & BM25               & 63.8            & 59.6           & 61.7         \\
\textbf{}                                                    & E5                 & 65.5            & 62.3           & 63.9         \\
\textbf{}                                                    & BGE                & 66.4            & 63.0           & 64.7         \\ \bottomrule
\end{tabular}
\caption{Performance comparison of different retrieval methods (BM25, E5, BGE) across models on the HotpotQA and 2WikiMultiHopQA datasets. Results show only the accuracy (ACC) metric .}
\label{tab:retriever_main}
\end{table}
We analyze the impact of different retrievers on our method compared with several baselines and show the results in Table~\ref{tab:retriever_main}.
Despite being trained exclusively with the E5 retriever, R3-RAG maintains its performance and consistently outperforms all baselines when using different retrievers, demonstrating the transferability of R3-RAG's learned reasoning and retrieval strategies across external retrieval environments.
Meanwhile, R3-RAG's performance variation across three different retrievers is less than 3\%, which further shows R3-RAG's robustness.
In contrast, baseline methods show significant fluctuation. For example, IRCoT's accuracy on the 2WikiMultiHopQA dataset plummets from 35.8\% when using E5 to just 20.1\% with BGE, revealing the human-designed workflow's sensitivity to different external retrieval environments. 
Additional results across more datasets and metrics are provided in Appendix~\ref{app:Retriever}.

\subsection{Efficiency Analysis}
\label{sec:efficiency}
\begin{table}[b]
\small
\centering
\begin{tabular}{@{}lccc@{}}
\toprule
\textbf{Method} & \textbf{2Wiki} & \textbf{MuSiQue} & \textbf{Avg} \\
\midrule
ReSearch & 506.97 & 541.97 & 524.46 \\
R3-RAG   & \textbf{382.08} & \textbf{429.94} & \textbf{405.01} \\
\midrule
Reduction (\%) $\downarrow$ & 24.6 & 20.7 & \textbf{22.8} \\
\bottomrule
\end{tabular}
\caption{Token usage comparison with \textbf{ReSearch} on the Qwen backbone. Values are average LLM tokens per correctly answered question across the multi-step interaction (lower is better).}
\label{tab:token-usage}
\end{table}
Longer responses that externalize reasoning often improve accuracy, but they also increase latency and token cost. 
To quantify this trade-off, we compare R3-RAG with \textbf{ReSearch}~\citep{chen2025researchlearningreasonsearch} on the Qwen backbone, reporting the average LLM token usage for correctly answered questions on 2WikiMultiHopQA and MuSiQue.
As shown in Table~\ref{tab:token-usage}, R3-RAG uses 22.8\% fewer tokens on average (24.6\% on 2WikiMultiHopQA; 20.7\% on MuSiQue) while maintaining higher accuracy (Table~\ref{tab:main-results}). 
We attribute this efficiency to our outcome-plus-process rewards: by jointly rewarding the final answer and each intermediate retrieval decision, R3-RAG retrieves more relevant documents and requires fewer, more productive reasoning turns than the outcome-only \textbf{ReSearch} baseline.

\subsection{Case Study} 
\begin{figure}[!h]
\centering
\includegraphics[width=0.47\textwidth]{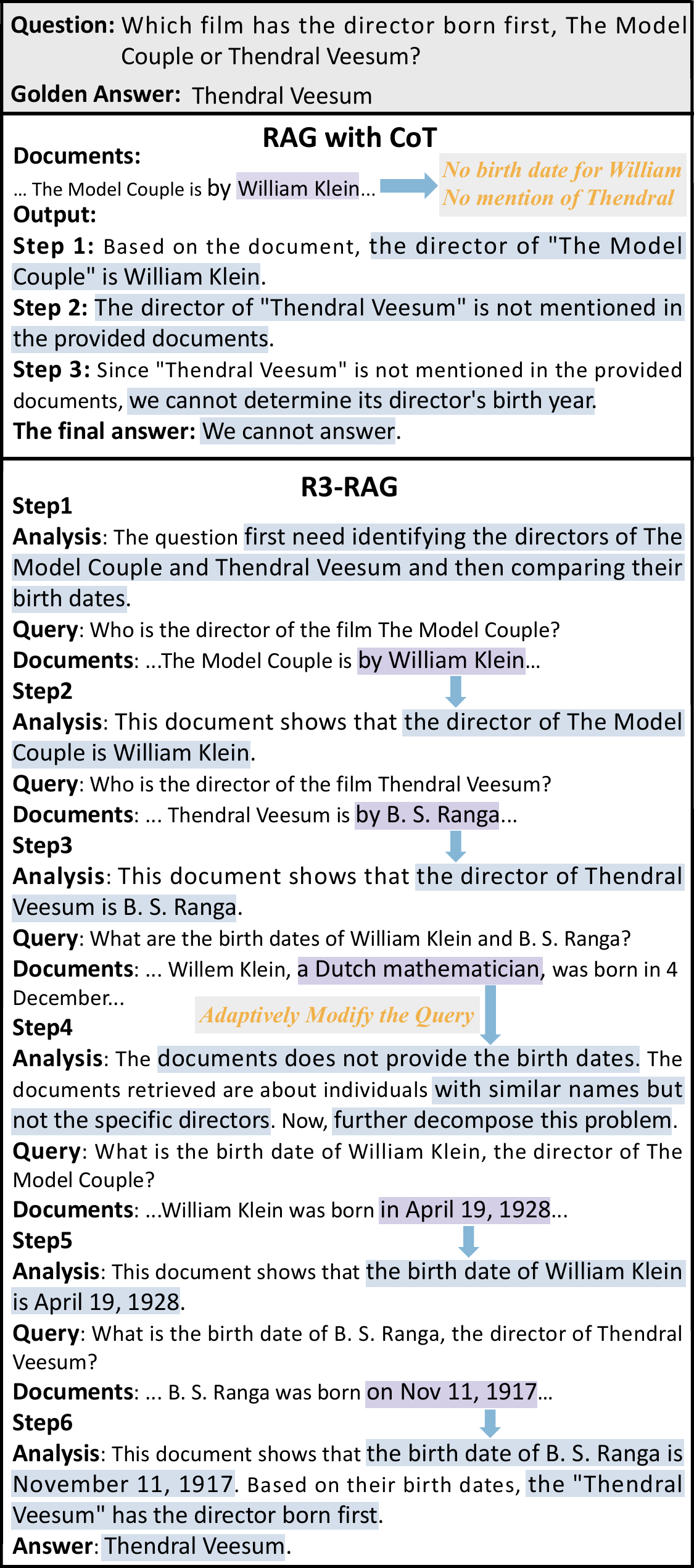}
\caption{Case Study: R3-RAG can reason and retrieve step-by-step and adaptively modify the query when retrieval fails.}
\label{fig:case}
\end{figure}

We conduct a case study on a multi-hop question (Figure~\ref{fig:case}) that asks for the film-The Model Couple or Thendral Veesum-whose director was born earlier. RAG + CoT fails to retrieve comprehensive information for this question. Specifically, it only retrieves the directors of these movies and thus fails to generate the answer. In contrast, R3-RAG can decompose the whole question into the following chain of retrieval queries: 1) the director of the film The Model Couple -> William Klein;
2) the director of the film Thendral Veesum -> B. S. Ranga; 3) the birth date of William Klein; and 4) the birth date of B. S. Ranga. Through iterative reasoning and retrieval, R3-RAG progressively retrieves relevant information and thus obtains comprehensive information for the question. Additionally, R3-RAG can adaptively decompose the previous query when it fails to retrieve relevant information. Specifically, when it tries to retrieve documents that contain both directors' birth dates and fails, it further decomposes this query into two separate queries and then successfully retrieves the corresponding information, which shows R3-RAG's potential when it encounters complex queries.

\section{Conclusion}
We propose R3-RAG, a framework that leverages reinforcement learning to teach LLMs how to reason and retrieve step-by-step, optimizing the interleaved reasoning and retrieval process with both outcome rewards and fine-grained process rewards through relevance-based document verification. Experimental results demonstrate that R3-RAG significantly outperforms existing iterative RAG methods across multiple benchmarks. Further analyses show the effectiveness of R3-RAG's reward design and its transferability across different retrievers.

\section*{Limitations}
Despite the effectiveness of R3-RAG, our approach has two main limitations. 
First, our experiments rely on publicly available academic datasets, which may lack sufficient diversity and realism, potentially limiting the model's robustness and generalization to real-world user queries. Future research could address this by curating more diverse and representative datasets.
Second, our cold-start data quality is constrained by the reasoning capabilities of the foundation model (DeepSeek-v3). Due to the model’s limitations, we could not reliably generate correct reasoning trajectories for certain challenging multi-hop questions, even after repeated sampling. Using stronger models, such as DeepSeek-R1, could further enhance cold-start data quality in future work.

\section*{Ethics Statement}
Our research complies with ethical standards in NLP research. We utilize publicly available datasets (HotpotQA, 2WikiMultiHopQA, and MuSiQue) and models (Llama-3.1-8B, Qwen2.5-7B) under appropriate licenses for academic research. Our derivative models comply with their original licenses. Detailed information about datasets, models, frameworks, and licensing can be found in Appendix~\ref{app:ethics_artifacts}.

Regarding data privacy and content safety, we conducted a thorough review of the datasets to ensure they do not contain personally identifiable information (PII) or offensive content. The datasets consist of questions and answers derived from Wikipedia, which undergoes content moderation. The questions focus on factual knowledge and do not target sensitive personal information. No additional data collection involving human subjects was conducted for this research, eliminating risks related to personal data exposure. Furthermore, we did not implement mechanisms that could generate harmful or offensive content. The training objectives were focused on improving factual accuracy and reasoning capabilities for answering multi-hop questions based on retrieved information.

\section*{Acknowledgements}
We would like to thank the anonymous reviewers for their insightful comments. 
This work was supported by the Shanghai Pilot Program for Basic Research - Fudan University 21TQ1400100 (22TQ018).

\bibliography{custom}

\clearpage

\appendix
\section{Online Data Sampling of RL}
\label{app:RLsampling}
Unlike the training of cold start or offline RLHF where data can be prepared in advance, PPO requires online sampling during the training process, with the dataset divided into multiple batches for sampling and training.

We use multi-hop question-answering datasets $\mathcal{Q}_{RL}$ that are distinct from the cold-start datasets ($\mathcal{Q}_{RL} \cap \mathcal{Q}_{IL} = \emptyset$) for reinforcement learning, with online sampling logic consistent with the cold-start phase. We denote the sampled dataset as $\mathcal{D}_{RL} = \{(q_i, a_i^*, T_i) \mid (q_i, a_i^*)\}$ and the final training dataset as $\mathcal{D}_{\mathrm{RL}}^{\mathrm{train}} = \{(\mathtt{input}_k, \mathtt{output}_k)\}_{k=1}^{m'}$, where $m' = \sum_{i=1}^{|\mathcal{Q}_{RL}|} |T_i|$, with each pair consisting of the question with previous reasoning steps as input and the current reasoning step as output.

A key distinction from the cold-start phase is that during reinforcement learning, we preserve all trajectories regardless of correctness or completeness. Consequently, the final step in any trajectory $T_i$ from $\mathcal{D}_{RL}$ may lack an answer or generate an incorrect answer.

\section{Generalized Advantage Estimation and KL Divergence Definition}
\label{app:gae}
The advantage function \( A(x, y) \) in reinforcement learning is used to guide policy improvement. We estimate \( A(x, y) \) using Generalized Advantage Estimation (GAE). In GAE, the advantage \( A(x, y) \) is defined as follows:
\begin{equation}
\small
A(x, y) = \sum_{t=0}^{T-1} (\gamma \lambda)^{t} \delta_t,
\end{equation}
where $ \delta_t $ is the Temporal Difference (TD) error at time step $ t $, which is $ \delta_t = r_t + \gamma V(y) - V(x) $. Specifically, $ x $ represents the current state $ s_t $, $ y $ represents the next state $ s_{t+1} $, $ r_t $ is the reward received at time step $ t $, $ \gamma $ is the discount factor, $ V(x) $ and $ V(y) $ are the value functions for states $ x $ and $ y $, respectively, and $ \lambda $ is a hyperparameter that controls the trade-off between bias and variance in the advantage estimation process.
The summation in the definition of $ A(x, y) $ captures the idea of taking into account not just the immediate rewards but also the future rewards, discounted over time, to provide a more accurate estimation of the advantage.

The term \(\mathcal{L}_{KL}\) represents a KL-divergence penalty to prevent the policy from deviating too much from the initial model:

\begin{equation}
\small
\mathcal{L}_{KL} = D_{KL}[P(y|x; \theta_{init}) \parallel P(y|x; \theta)],
\end{equation}
where \(\beta\) controls the strength of this regularization. The details of the process reward model will be provided in the next section.

\section{Implementation Details}
\label{app:details}
We use Llama-3.1-8B \cite{grattafiori2024llama3herdmodels} and Qwen2.5-7B \cite{qwen2.5} models for training and evaluating. 
For the stage of the cold start, we prompt DeepSeek-V3 \cite{deepseekai2024deepseekv3technicalreport} models to obtain step-by-step solution trajectories that can solve problems correctly with a specific format. We sampled from 15,000 examples in the 2WikiMultiHopQA training set, 19,937 examples in the MuSiQue training set, and the first 30,000 examples in HotpotQA, resulting in 10,843, 14,145, and 26,266 solution trajectories respectively. 
Then we further cleaned these data to ensure they conform to our required format specifications. We obtained a total of 51,254 trajectories for cold start, which can arrive at correct answers but do not necessarily follow optimal solution paths.
We use these lower-quality data to train our base model through the LLaMA-Factory framework \cite{zheng2024llamafactory}, resulting in the first version of our method, which we call R3-RAG-CS. During training, we apply full-parameter fine-tuning with DeepSpeed zero-3 optimization and a maximum token length of 4,096. For Llama and Qwen models, we set learning rates at 1.0e-5 and 7.0e-6, with batch sizes of 8 and 16 respectively.

For the stage of RL, since we have fully utilized the 2WikiMultiHopQA and MuSiQue training datasets, we extract 8,192 samples from the remaining 60,000+ HotpotQA training data for reinforcement learning. We modified the sampling code in the OpenRLHF framework \cite{hu2024openrlhf} to enable step-by-step sampling and tool calling according to our requirements. Then we use the modified OpenRLHF to complete PPO reinforcement learning training, resulting in our R3-RAG models. Actor learning rates are set at 5e-7 for both models, critic learning rates at 9e-6, and the training batch size is uniformly set at 64. The context length after multiple inference iterations and model calls may lead to CUDA out of memory errors. To avoid frequent errors, we limit each retrieved document to 512 characters during reinforcement learning training. we set $\gamma_{fail} = -0.1$, $\gamma_{incorrect} = -0.2$, and $\gamma_{correct} = 0.6$. And we use Qwen2.5-14B-Instruct \cite{qwen2.5} to assign scores of document relevance.

We use the wikipedia-nq-corpusfrom Tevatron (wikipedia2018) with the FlashRAG \cite{FlashRAG} framework to build the retriever. If the retriever is not specified, we default to using the E5-base-v2 model\cite{wang2022text} as the retriever. Additionally, for fair testing, we also provide results using 
BGE-large-en-v1.5\cite{bge_embedding} and BM25\cite{bm25s}. When using the E5 and BGE models, We use the faiss library \cite{johnson2019billion} for GPU-accelerated retrieval.

In evaluation, although we know that excessive context length can degrade model reasoning performance, we directly input the retrieved content into the model without extracting key information as other methods do. This better represents the model's capabilities in real-world environments, where useful information is often hidden among large amounts of irrelevant information.
We set the retriever's top-$k$ to 5 and limit the maximum number of iterations to 5 during evaluation. Specifically, for the Naive RAG method, since each question only retrieves once, we set the number of top-k to 16, as our proposed method uses fewer than 16 documents on average. If the model fails to answer a question within this number of steps, we consider it unable to answer the question and count it as incorrect. We implemented Naive Generation, Naive RAG, and react methods, prompting the models to generate answers step by step. For baselines including Flare, Self-ask, ITER-RETGEN, and IRCoT, we used the FlashRAG implementation. All methods first generate answers, which are then evaluated using a unified evaluation script. Evaluation data for Auto-RAG and DeepRAG are cited from the DeepRAG\cite{guan2025deepragthinkingretrievalstep}, while ReSearch-Qwen-7B-Instruct results are cited from its paper\cite{chen2025researchlearningreasonsearch}. We ensured consistent evaluation parameters across all methods. And we use Qwen2.5-72B-Instruct \cite{qwen2.5} to judge the correctness for the accuracy in our metrics.

We conducted all our model training experiments twice to ensure the training process is reproducible. During evaluation, we observe that the model's output results were stable, therefore we only performed a single trial. We conduct our experiments using NVIDIA H100 GPUs.

\section{Details of Analysis for Retriever Quality}
\label{app:Retriever}
\begin{table*}[t]
\small
\centering
\begin{tabular}{@{}
>{\columncolor[HTML]{FFFFFF}}l 
>{\columncolor[HTML]{FFFFFF}}c 
>{\columncolor[HTML]{FFFFFF}}l 
>{\columncolor[HTML]{FFFFFF}}l 
>{\columncolor[HTML]{FFFFFF}}l 
>{\columncolor[HTML]{FFFFFF}}l 
>{\columncolor[HTML]{FFFFFF}}l 
>{\columncolor[HTML]{FFFFFF}}l 
>{\columncolor[HTML]{FFFFFF}}l 
>{\columncolor[HTML]{FFFFFF}}l 
>{\columncolor[HTML]{FFFFFF}}l 
>{\columncolor[HTML]{FFFFFF}}l @{}}
\toprule
\multicolumn{1}{c}{\cellcolor[HTML]{FFFFFF}}                                   & \cellcolor[HTML]{FFFFFF}                                     & \multicolumn{3}{c}{\cellcolor[HTML]{FFFFFF}\textbf{HotpotQA}}                                                                                     & \multicolumn{3}{c}{\cellcolor[HTML]{FFFFFF}\textbf{2WikiMultiHopQA}}                                                                              & \multicolumn{3}{c}{\cellcolor[HTML]{FFFFFF}\textbf{MuSiQue}}                                                                                      & \multicolumn{1}{c}{\cellcolor[HTML]{FFFFFF}\textbf{Avg}} \\ \cmidrule(l){3-12} 
\multicolumn{1}{c}{\multirow{-2}{*}{\cellcolor[HTML]{FFFFFF}\textbf{Methods}}} & \multirow{-2}{*}{\cellcolor[HTML]{FFFFFF}\textbf{Retriever}} & \multicolumn{1}{c}{\cellcolor[HTML]{FFFFFF}ACC} & \multicolumn{1}{c}{\cellcolor[HTML]{FFFFFF}F1} & \multicolumn{1}{c}{\cellcolor[HTML]{FFFFFF}EM} & \multicolumn{1}{c}{\cellcolor[HTML]{FFFFFF}ACC} & \multicolumn{1}{c}{\cellcolor[HTML]{FFFFFF}F1} & \multicolumn{1}{c}{\cellcolor[HTML]{FFFFFF}EM} & \multicolumn{1}{c}{\cellcolor[HTML]{FFFFFF}ACC} & \multicolumn{1}{c}{\cellcolor[HTML]{FFFFFF}F1} & \multicolumn{1}{c}{\cellcolor[HTML]{FFFFFF}EM} & \multicolumn{1}{c}{\cellcolor[HTML]{FFFFFF}ACC}          \\ \midrule
\multicolumn{12}{c}{\cellcolor[HTML]{EFEFEF}\textbf{Llama-3.1-8B}}                                                                                                                                                                                                                                                                                                                                                                                                                                                                                                                                                                                                   \\
RAG with CoT                                                                   & BM25                                                         & 52.3                                            & 49.9                                           & 38.1                                           & 31.5                                            & 39.4                                           & 30.5                                           & 13.3                                            & 16.3                                           & 8.0                                            & 32.4                                                     \\
\textbf{}                                                                      & E5                                                           & 53.3                                            & 50.8                                           & 38.6                                           & 32.9                                            & 40.2                                           & 31.1                                           & 16.3                                            & 18.8                                           & 10.3                                           & 34.2                                                     \\
\textbf{}                                                                      & BGE                                                          & 55.8                                            & 52.8                                           & 40.5                                           & 33.9                                            & 41.1                                           & 32.4                                           & 16.8                                            & 19.5                                           & 10.3                                           & 35.5                                                     \\
ReAct(Tool Call)                                                               & BM25                                                         & 33.6                                            & 28.3                                           & 23.3                                           & 18.0                                            & 15.4                                           & 13.6                                           & 8.4                                             & 5.8                                            & 4.3                                            & 20.0                                                     \\
                                                                               & E5                                                           & 30.8                                            & 25.8                                           & 21.2                                           & 17.8                                            & 14.5                                           & 12.9                                           & 8.6                                             & 6.5                                            & 4.8                                            & 19.1                                                     \\
                                                                               & BGE                                                          & 29.8                                            & 25.2                                           & 20.6                                           & 16.1                                            & 13.2                                           & 11.6                                           & 8.2                                             & 6.0                                            & 4.6                                            & 18.0                                                     \\
IRCoT                                                                          & BM25                                                         & 52.8                                            & 46.4                                           & 39.5                                           & 41.4                                            & 38.6                                           & 36.2                                           & 17.5                                            & 14.0                                           & 12.0                                           & 37.2                                                     \\
                                                                               & E5                                                           & 52.8                                            & 46.0                                           & 39.3                                           & 40.6                                            & 37.5                                           & 35.1                                           & 16.7                                            & 13.6                                           & 12.0                                           & 36.7                                                     \\
                                                                               & BGE                                                          & 43.3                                            & 37.8                                           & 32.4                                           & 20.8                                            & 18.9                                           & 17.9                                           & 12.8                                            & 10.3                                           & 9.0                                            & 25.6                                                     \\
\midrule[0.03em]
R3-RAG-CS                                                                      & BM25                                                         & 57.8                                            & 53.3                                           & 41.6                                           & 48.0                                            & 49.1                                           & 42.3                                           & 22.6                                            & 23.0                                           & 15.4                                           & 42.8                                                     \\
                                                                               & E5                                                           & 60.6                                            & 55.5                                           & 43.6                                           & 53.7                                            & 53.4                                           & 46.5                                           & 29.6                                            & 29.4                                           & 19.4                                           & 48.0                                                     \\
                                                                               & BGE                                                          & 61.7                                            & 56.8                                           & 44.5                                           & 54.6                                            & 54.4                                           & 47.4                                           & 28.8                                            & 28.9                                           & 19.4                                           & 48.4                                                     \\
R3-RAG                                                                         & BM25                                                         & 62.5                                            & 57.6                                           & 44.4                                           & 58.0                                            & 58.6                                           & 50.6                                           & 26.4                                            & 27.7                                           & 17.2                                           & 49.0                                                     \\
\textbf{}                                                                      & E5                                                           & 64.4                                            & 58.8                                           & 45.6                                           & 61.0                                            & 60.9                                           & 52.9                                           & 32.2                                            & 32.7                                           & 21.1                                           & 52.5                                                     \\
\textbf{}                                                                      & BGE                                                          & 65.3                                            & 60.0                                           & 46.6                                           & 62.1                                            & 61.8                                           & 53.7                                           & 33.8                                            & 32.8                                           & 21.2                                           & 53.7                                                     \\ \midrule
\multicolumn{12}{c}{\cellcolor[HTML]{EFEFEF}\textbf{Qwen2.5-7B}}                                                                                                                                                                                                                                                                                                                                                                                                                                                                                                                                                                                                     \\
RAG with CoT                                                                   & BM25                                                         & 49.9                                            & 47.8                                           & 36.0                                           & 29.8                                            & 36.4                                           & 27.4                                           & 14.3                                            & 16.8                                           & 8.7                                            & 31.3                                                     \\
                                                                               & E5                                                           & 52.4                                            & 49.4                                           & 37.6                                           & 33.5                                            & 39.1                                           & 30.1                                           & 16.9                                            & 18.8                                           & 9.9                                            & 34.3                                                     \\
                                                                               & BGE                                                          & 55.1                                            & 51.7                                           & 39.6                                           & 35.3                                            & 40.5                                           & 32.0                                           & 18.3                                            & 20.9                                           & 11.3                                           & 36.2                                                     \\
ReAct(Tool Call)                                                               & BM25                                                         & 43.4                                            & 36.0                                           & 29.4                                           & 28.7                                            & 25.1                                           & 23.0                                           & 11.3                                            & 7.6                                            & 5.5                                            & 27.8                                                     \\
                                                                               & E5                                                           & 27.9                                            & 23.1                                           & 21.1                                           & 38.3                                            & 31.7                                           & 26.1                                           & 11.7                                            & 8.0                                            & 5.8                                            & 26.0                                                     \\
                                                                               & BGE                                                          & 37.7                                            & 31.4                                           & 26.0                                           & 27.4                                            & 22.6                                           & 20.6                                           & 11.9                                            & 8.2                                            & 5.8                                            & 25.7                                                     \\
IRCoT                                                                          & BM25                                                         & 48.5                                            & 42.4                                           & 36.1                                           & 33.8                                            & 31.7                                           & 29.6                                           & 12.8                                            & 10.7                                           & 9.1                                            & 31.7                                                     \\
                                                                               & E5                                                           & 48.4                                            & 42.1                                           & 35.7                                           & 35.8                                            & 33.5                                           & 31.1                                           & 13.5                                            & 11.2                                           & 9.4                                            & 32.6                                                     \\
                                                                               & BGE                                                          & 40.5                                            & 35.3                                           & 30.0                                           & 20.1                                            & 18.9                                           & 18.2                                           & 9.2                                             & 7.7                                            & 6.5                                            & 23.3                                                     \\
\midrule[0.03em]
R3-RAG-CS                                                                      & BM25                                                         & 60.4                                            & 55.5                                           & 43.5                                           & 49.8                                            & 50.4                                           & 43.5                                           & 25.1                                            & 25.3                                           & 16.2                                           & 45.1                                                     \\
                                                                               & E5                                                           & 63.3                                            & 57.6                                           & 45.2                                           & 53.1                                            & 53.5                                           & 46.1                                           & 31.3                                            & 31.9                                           & 21.9                                           & 49.2                                                     \\
                                                                               & BGE                                                          & 64.1                                            & 58.7                                           & 46.1                                           & 55.2                                            & 55.2                                           & 48.1                                           & 31.1                                            & 31.1                                           & 21.1                                           & 50.1                                                     \\
R3-RAG                                                                         & BM25                                                         & 63.8                                            & 58.2                                           & 44.9                                           & 59.6                                            & 61.1                                           & 52.8                                           & 29.2                                            & 30.0                                           & 17.6                                           & 50.9                                                     \\
\textbf{}                                                                      & E5                                                           & 65.5                                            & 59.7                                           & 46.4                                           & 62.3                                            & 62.7                                           & 54.2                                           & 33.6                                            & 34.0                                           & 21.4                                           & 53.8                                                     \\
\textbf{}                                                                      & BGE                                                          & 66.4                                            & 60.6                                           & 46.8                                           & 63.0                                            & 63.4                                           & 55.2                                           & 34.8                                            & 34.3                                           & 21.7                                           & 54.7                                                     \\ \bottomrule
\end{tabular}
\caption{Comprehensive performance comparison of different retrieval methods (BM25, E5-base-v2, BGE-Large) across all models on HotpotQA, 2WikiMultiHopQA, and MuSiQue datasets. Results include accuracy (ACC), F1, and exact match (EM) metrics.}
\label{tab:analysis-retriever}
\end{table*}
We present a comprehensive evaluation of datasets and metrics in Table~\ref{tab:analysis-retriever}. The results show that our model consistently outperforms all baselines across all metrics on the three datasets. Moreover, R3RAG consistently surpasses R3-RAG-CS, further demonstrating the transferability of R3-RAG's learned reasoning and retrieval strategies across external retrieval environments.

Interestingly, among the baseline methods, both models struggle to effectively utilize the stronger BGE retriever, often performing better with the theoretically weaker BM25 and E5 retrievers. 
We propose two explanations for this counter-intuitive phenomenon. First, different models have varying adaptation capabilities to retriever characteristics, meaning that stronger general-purpose retrievers don't always provide better document recall for all models and baseline query formulations. Second, our analysis of retrieval queries and returned documents reveals that baseline methods typically generate broad queries containing multiple knowledge points, expecting retrievers to return comprehensive information in a single operation. However, these knowledge points often have low semantic similarity to each other (e.g., different entity names). BM25 and E5, especially BM25, tend to prioritize documents with high relevance to individual knowledge points, while BGE attempts to optimize for average relevance across all query components, potentially resulting in lower relevance for each specific knowledge point and consequently fewer critical documents compared to BM25.

\section{Potential Risk}
Previous works have shown LLMs can have various kinds of bias\cite{llm_bias}. Since our method uses training data distilled from DeepSeek-v3 during the cold-start phase, it can also inherit such biases.

\section{Details of Research Artifacts and Licenses}
\label{app:ethics_artifacts}
This research utilizes several publicly available datasets, pre-trained models, and frameworks. We provide comprehensive details on their licenses and our usage to ensure transparency and compliance with ethical research standards.

\subsection{Datasets and Licenses}
\label{app:datasets_licenses}
We utilized three multi-hop QA datasets:
\begin{itemize}
    \item HotpotQA \cite{yang2018hotpotqadatasetdiverseexplainable} - Distributed under the CC BY-SA 4.0 license
    \item 2WikiMultiHopQA \cite{ho2020constructingmultihopqadataset} - Distributed under the Apache-2.0 license
    \item MuSiQue \cite{trivedi2022musiquemultihopquestionssinglehop} - Distributed under the CC BY 4.0 license
\end{itemize}

Our usage of these datasets adheres strictly to their intended research purposes as specified by their creators. All datasets were used in their original form for academic research purposes only, and no modifications were made to their licenses. The datasets contain questions derived from Wikipedia content and are designed specifically for multi-hop question answering research.

For the artifacts we created in this study (fine-tuned R3-RAG-CS and R3-RAG models), we maintain the same licensing terms as their base models and specify that they are intended for academic research purposes only. These derivative models are compatible with the original access conditions of their base models and datasets, and should not be used outside of research contexts. Our modifications were limited to parameter fine-tuning and did not alter the fundamental architecture or intended use cases of the original models.

\subsection{Models, Frameworks, and Their Licenses}
\label{app:models_frameworks}
We employed the following models and frameworks:

\textbf{Language Models:}
\begin{itemize}
    \item Llama-3.1-8B \cite{grattafiori2024llama3herdmodels} - Available under the Llama 3.1 Community License
    \item Qwen2.5-7B \cite{qwen2.5} - Available under the Apache-2.0 license
\end{itemize}

\textbf{Retriever:}
\begin{itemize}
    \item E5-base-v2 \cite{wang2022text} - Available under the MIT license
    \item BGE-large-en-v1.5 \cite{bge_embedding} - Available under the MIT license
    \item BM25S \cite{bm25s} - Available under the MIT license
\end{itemize}

\textbf{Frameworks:}
\begin{itemize}
    \item LLaMA-Factory \cite{zheng2024llamafactory} - Available under the Apache-2.0 license
    \item OpenRLHF \cite{hu2024openrlhf} - Available under the Apache-2.0 license
    \item FlashRAG \cite{FlashRAG} - Available under the MIT license
    \item FAISS \cite{johnson2019billion} - Available under the MIT license
\end{itemize}

Our use of these artifacts is consistent with their intended purposes. The language models and retrieval models were used for natural language processing and information retrieval tasks within a research context. The frameworks were employed as development tools to implement our research methodology. No commercial applications were developed using these resources. All model adaptations created during this research maintain compatibility with the original license terms of their respective base models.

\begin{figure*}[t]
\begin{lstlisting}
def prompt_question_init(question):
   return f'''
# Role
You are an expert in large language models and knowledge retrieval, with extensive expertise in problem decomposition.
# Answer Format:
The problem analysis: [Analyze information based on the question.]
**Either:**
The retrieval query: [If information is insufficient, generate one query.]
**OR:**
The final answer: [If information is sufficient, provide the final answer.]
# Instructions:
1. **The Problem Analysis**:
   - **Decision Making**:
     - **Sufficient Information**: If the available information is adequate and no more query need to retrieve, use your parameter knowledge and reasoning capability to solve it and get the answer. Finaly, output the answer in the "The final answer:" section.
     - **Insufficient Information**: If the information is lacking, identify what information is missing:
       - **Determine Next Retrieval**: Consider what content should be retrieved next and decomposite the query into simpler retrieval queries.
2. **Response Formatting**:
   - **Start with Analysis**: Always begin with "The problem analysis:" followed by the detailed analysis. Only one analysis is permitted; therefore, include all your analytical content within this section.
   - **Choose Appropriate Section**: Based on the decision:
     - Include "The retrieval query:" if more information is needed.
     - Include "The final answer:" if sufficient information is available.
   - **Generate a Single Query**: Based on the analysis, formulate one targeted query and place it in the "The retrieval query:" section. Ensure that only one query is generated, focusing on the first retrieval question needed.
   - **Language Consistency**: Provide the entire response in English, regardless of the input language.
# Key Competencies
1. **Decision Making**:
   - **Assessment**: Evaluate whether it is feasible to resolve the entire problem based on all preceding steps.
   - **Action in Uncertainty**: If uncertain, initiate an attempt to solve the problem.
   - **Failure Handling**: If the solution process encounters a failure, terminate the attempt and analyze which information should be retrieved next.
2. **Decomposition of Retrieval Content**:
   - **Problem Breakdown**: When analysis requires solving a problem, decompose complex questions into smaller, actionable sub-questions (atomic problems).
   - **Step-by-Step Resolution**: Address these sub-questions sequentially and establish a logical sequence to answer the original question.
   - **Identifying Broad Problems**: If the analysis reveals that the information retrieved in the last step does not align with the query, this indicates that the problem is too broad. In such cases:
     - **Suggestion**: Propose a method to decompose the problem into smaller sub-questions.
     - **Implementation**: Attempt to break down the original query into simpler, more manageable questions.
   - **Methods of Decomposition**: Complex questions can be decomposed into **two fundamental forms**:
     - **Sequential Decomposition**:
       - **Definition**: Break the problem into a series of dependent steps, where each step relies on the result of the previous one.
     - **Parallel Decomposition**:
       - **Definition**: Split the problem into multiple independent sub-problems, solve each separately, and then combine the results.
#### Current Input: The question: {question}
#### Output:
'''
\end{lstlisting}
\caption{The initial prompt template used for the first step of trajectory generation.}
\label{fig:template1}
\end{figure*}

\begin{figure*}[t]
\begin{lstlisting}
def prompt_question(steps_formatted, question):
   return f'''
# Role: You are an expert in large language models and knowledge retrieval.
# Answer Format:
The problem analysis: [Analyze information based on previous steps and the question]
**Either:**
The retrieval query: [If information is insufficient, generate one query]
**OR:**
The final answer: [If information is sufficient, provide the final answer]
# Instructions:
1. **The Problem Analysis**:
   - **Review Previous Steps**: Examine all prior steps, including analyses, queries, and documents.
   - **Decision Making**:
     - **Sufficient Information**: If the information is adequate, analyze it to solve the problem and output the answer in the "The final answer:" section.
     - **Insufficient Information**: If the information is lacking:
       - **Evaluate Last Step**: Analyze the documents obtained in the last step to determine if they are relevant to the query or the problem at hand.
       - **Utilizing parameter knowledge**: If the retrieval results of the query are not relevant, check if your parameter knowledge can response the query.
       - **Assess Previous Steps**: Identify what information is missing from the overall previous steps.
       - **Determine Next Retrieval**: Consider what content should be retrieved next and decomposite the query into simpler retrieval queries.
2. **Response Formatting**:
   - **Start with Analysis**: Always begin with "The problem analysis:" followed by the detailed analysis. Only one analysis is permitted; therefore, include all your analytical content within this section.
   - **Choose Appropriate Section**: Based on the decision:
     - Include "The retrieval query:" if more information is needed.
     - Include "The final answer:" if sufficient information is available.
   - **Generate a Single Query**: Based on the analysis, formulate one targeted query and place it in the "The retrieval query:" section. Ensure that only one query is generated, focusing on the first retrieval question needed.
   - **Language Consistency**: Provide the entire response in English, regardless of the input language.
# Key Competencies
1. **Decision Making**:
   - **Assessment**: Evaluate whether it is feasible to resolve the entire problem based on all preceding steps.
   - **Action in Uncertainty**: If uncertain, attempt to solve the problem.
   - **Failure Handling**: If the solution process encounters a failure, terminate the attempt and analyze which information should be retrieved next.
2. **Decomposition of Retrieval Content**:
   - **Problem Breakdown**: When analysis requires solving a problem, decompose complex questions into smaller, actionable sub-questions (atomic problems).
   - **Step-by-Step Resolution**: Address these sub-questions sequentially and establish a logical sequence to answer the original question.
   - **Identifying Broad Problems**: If the analysis reveals that the information retrieved in the last step does not align with the query, this indicates that the problem is too broad. In such cases:
     - **Suggestion**: Propose a method to decompose the problem into sub-questions.
     - **Implementation**: Attempt to break down the original query into simpler, more manageable questions.
   - **Methods of Decomposition**: Complex questions can be decomposed into **two fundamental forms**:
     - **Sequential Decomposition**:
       - **Definition**: Break the problem into a series of dependent steps, where each step relies on the result of the previous one.
     - **Parallel Decomposition**:
       - **Definition**: Split the problem into multiple independent sub-problems, solve each separately, and then combine the results.
#### Current Input: The question: {question}. The Previous Steps:{steps_formatted}
#### Output:
'''
\end{lstlisting}
\caption{The prompt template used for subsequent steps in trajectory generation.}
\label{fig:template2}
\end{figure*}

\end{document}